\pdfoutput=1

\documentclass[11pt]{article}

\usepackage{EACL2023}

\usepackage{times}
\usepackage{latexsym}

\usepackage[T1]{fontenc}

\usepackage[utf8]{inputenc}

\usepackage{microtype}

\usepackage{inconsolata}
\usepackage{graphicx}
\usepackage{amsmath}
\usepackage{multirow}
\usepackage{adjustbox}
\usepackage{booktabs}
\usepackage{longtable}
\usepackage{subfig}
\newcommand*{\img}[1]{%
    \raisebox{-.3\baselineskip}{%
        \includegraphics[
        height=\baselineskip,
        width=\baselineskip,
        keepaspectratio,
        ]{#1}%
    }%
}

\title{VOLTAGE: A Versatile Contrastive Learning based OCR Methodology for ultra low-resource scripts through Auto Glyph Feature Extraction}

\author{Prawaal Sharma \\
  Infosys\\
  Pune, Maharashtra, India \\ 
  prawaal\_sharma@infosys.com\\ 
  \And
  Poonam Goyal \\
  BITS Pilani \\
  Pilani, Rajasthan, India \\
  poonam@pilani.bits-pilani.ac.in\\
  \AND
  Vidisha Sharma \\
  Pune, Maharashtra, India \\
  vidishasharma@gmail.com\\
  \And
  Navneet Goyal \\
  BITS Pilani \\
  Pilani, Rajasthan, India \\
  goel@pilani.bits-pilani.ac.in\\
  } 

\begin{document}
\maketitle
\begin{abstract}
UNESCO has classified 2500 out of 7000 languages spoken worldwide as endangered. Attrition of a language leads to loss of traditional wisdom, folk literature, and the essence of the community that uses it.  It is therefore imperative to bring digital inclusion to these languages and avoid its extinction. Low resource languages are at a greater risk of extinction. Lack of unsupervised Optical Character Recognition(OCR) methodologies for low resource languages is one of the reasons impeding their digital inclusion. We propose VOLTAGE - a contrastive learning based OCR methodology, leveraging auto-glyph feature recommendation for cluster-based labelling. We augment the labelled data for diversity and volume using image transformations and Generative Adversarial Networks. Voltage has been designed using Takri - a family of scripts used in 16th to 20th century in the Himalayan regions of India. We present results for Takri along with other Indic scripts (both low and high resource) to substantiate the universal behavior of the methodology. An accuracy of 95\% for machine printed and 87\% for handwritten samples on Takri script has been achieved. We conduct baseline and ablation studies along with building downstream use cases for Takri, demonstrating the usefulness of our work.

\end{abstract}
\begin{figure*}
\centering
{
\fbox{\includegraphics[width=0.95\textwidth]{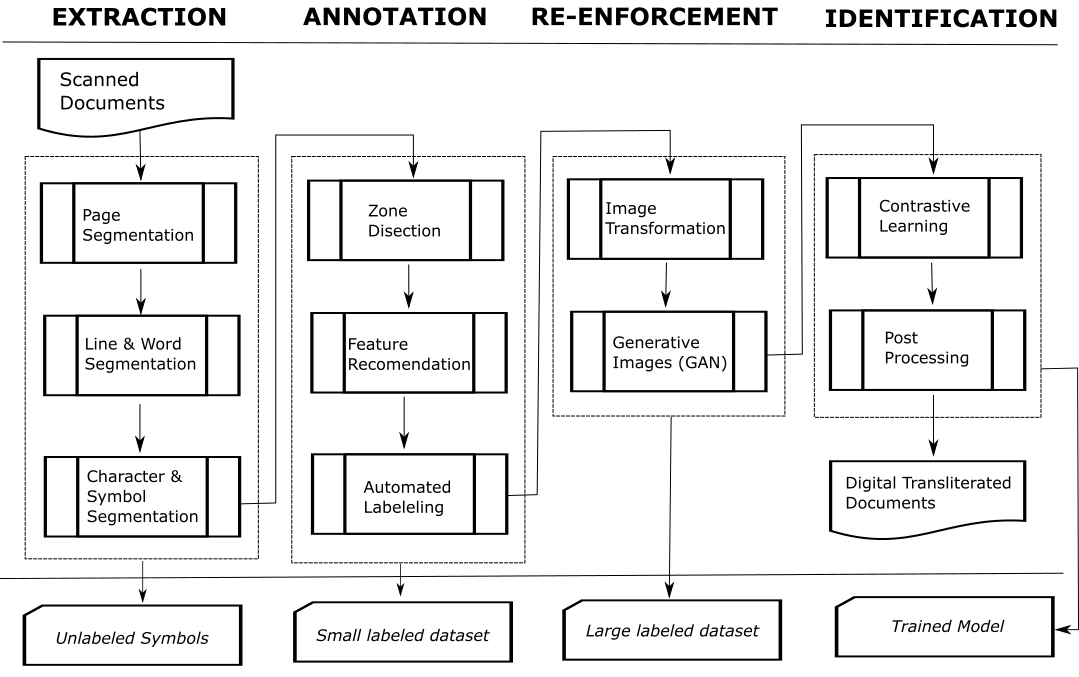}}
}
\setlength{\belowcaptionskip}{-15pt}  
\caption{High level design for \textbf{V}ersatile unsupervised \textbf{O}CR methodology for \textbf{L}ow-resource scripts \textbf{T}hrough \textbf{A}uto \textbf{G}lyph feature \textbf{E}xtraction (VOLTAGE)}
\label{Architecture}
\end{figure*}

\section{Introduction}
The UNESCO "Atlas of the World's Languages in Danger" \cite{unesco} is considered as a benchmark for the comprehensive list of the world's endangered languages. This study unveils more than 2500 languages and dialects as endangered out of which 200 come from Indian demography.

Optical character recognition (OCR) is used for digitizing historical archives, helping language conservation. There are plenty commercial and open-source OCR engines available for contemporary documents. However, very Low Resource Scripts (LRS) differ in their requirements mainly because of non-availability of large volume of data and limited users. The two most popular unsupervised (or semi-supervised) OCR methods available include Ocular \cite{ocular} and anyOCR \cite{anyocr}. Both methods are designed for large datasets and hence cannot be applied to LRS effectively.

Another alternative is to apply pretrained models for a high resource language as a foundation model and apply few shot learning to customize it and get the desired results. Our experiments conducted for "Takri" using this approach do not result in good accuracy. We have discussed this in detail in results section.

We develop an automated versatile unsupervised OCR methodology (VOLTAGE) for very low resource scripts to address the gap. We use Takri as an example to develop our methodology due to (a) No available labelled data and scanty user base (b) Extremely low digital unlabelled resources. We further evaluate the proposed methodology on four other languages to validate the universal behavior.

As illustrated in Figure \ref{Architecture}, VOLTAGE comprises of four steps, (a) Extraction: segmentation of available data into pages, lines, words, characters and symbols; (b) Annotation: feature extraction and recommendation followed by cluster based labelling; (c) Re-enforcement: augmentation of dataset using image transformation and generative AI (GANs); (d) Identification: contrastive learning based classification for character identification. The novelty of the methodology is that the manual intervention including human oracles is bare-minimum. The proposed glyph pattern-based feature recommender system can be applied to any script to recommend apropos feature set.

We empirically discuss our results in detail for Takri and also evaluate on other Indic scripts to validate its generalization capabilities. We also conduct baseline and ablation studies to substantiate our results. The contribution summary of our work is three fold:

\begin{itemize}
\item We build versatile and automated OCR methodology using contrastive learning approach for ultra low resource scripts.
\item We build a novel glyph feature recomendor system for unsupervised labelling of symbols which can be applied universally.
\item We build the largest labelled Takri dataset containing approximately 226,000 symbols, along with downstream use cases on transliteration and synthetic symbol generation models for public use. 
 \end{itemize}

\section{Related Work}
\subsection{Optical Character Recognition (OCR)}
The early ideas of OCR dates back to 1870's \cite{ocrHistory}. Since then, the OCR systems have evolved, and in the modern world there are many open source OCR systems like Tessaract \cite{textract}, OCRopus \cite{ocropus}, Kraken \cite{Kraken} and Calamari \cite{calamari} etc. Although research on OCR for Indic scripts started only in the mid 1970s, however the scope of research was restricted to Devanagari, Tamil and Telugu scripts only \cite{oldOCRIndic}. Even today, the major work in Indic OCR is limited to the ten scripts namely, Bangla, Devanagari, Gurumukhi, Gujarati, Kannada, Malayalam, Oriya, Tamil, Telugu, and Urdu \cite{IndicOCR}. Most of the current OCRs are based on deep neural networks which tends to be hungry on data and computational power. 

OCR pipeline generally goes through multiple individual tasks including (a) Image acquisition (extracting images containing text from multiple sources for offline images, and capturing live images for online extraction) (b) Pre-processing (application of image processing techniques, to increase raw image quality)  (b) Binarization (for scenarios where text and images/videos are mixed, we need to isolate text images from background) (c) Layout Analysis (dividing the images into regions) (d) Segmentation (segmentation of image into pages, lines, words, characters and symbols)  (e) Feature Analysis (identification and extraction of key features) (f) Classification (Recognition of symbol with scrip character-set)  (g) Post processing (use of pre-compiled vocabulary and language rules to auto correct the unrecognized words) \cite{ocrprocess}.

\textbf{Supervised OCR:} uses labelled dataset for training the classifier. Supervised methods give better performance however, annotation of character level images needs a lot of efforts and is not practical for low resource languages (LRL) where availability of annotators is scarce. Most supervised SOTA OCR systems like Tessarct and OCRopus are pre-trained on a very large image data sets based on deep CNN neural networks \cite{cnn}.

\textbf{Unsupervised OCR:} Unsupervised transformers like BERT \cite{bert}, GPT \cite{gpt} etc. have become very successful for diverse NLP tasks. In case of OCR systems only a few unsupervised (or semi-supervised) methods are available like Ocular \cite{ocular} and anyOCR \cite{anyocr}. Ocular uses generative modelling approach incorporating font typesetting, inking and noise. AnyOCR on the other hand is semi-supervised and language agnostic which consumes historical documents and clusters them for training purpose.

From the best of our knowledge, there is no general purpose OCR methodology suitable for ultra low resource scripts used by very limited user groups, with limited or no digital data available. VOLTAGE fills that gap, with the design of autonomous OCR pipeline enabling digitization of ultra low resource scripts.

\subsection{Data Augmentation}
Data Augmentation (DA) helps with increase of volume and diversity of data. With the advent of deep learning methods where the efficiency and accuracy of models is proportional to the training data, it has become imperative to use data augmentation approaches to generate large volumes of synthetic data and improve the performance of the model \cite{ocrSynthetic}.

Data augmentation for images is classified into two categories (a) Extractive, which augments data applying rules and transformations in form of rotation, brightness, sheer, zoom, flips etc. \cite{extractive1, extractive2} and (b) Generative, which synthesises data based on existing patterns using Generative Adversarial Networks (GAN) \cite{gan}. Generative methods helps in expanding the diversity of textual images along with inclusion of noise, and is therefore is very close to human generated samples \cite{gan1, gan2, gan3}.

\subsection{Contrastive Learning}
The use of contrastive learning in various NLP and computer vision tasks is becoming very popular in recent years \cite{contrastive1}, including sentence embeddings \cite{gao}, language translation \cite{pan}, text generation \cite{shu} etc. Contrastive learning can be applied in a self supervised mode, where the anchor and the positive sample are pulled together in embedding space, the negative samples are pushed apart \cite{chen, tian}. Another approach of using contrastive learning is in supervised mode where multiple positives per anchor are pulled closer together along with many negatives anchors which are pulled further \cite{SupCon}. The contrastive losses in this case is the generalization of triplet \cite{triplet} and N-Pair \cite{n-pair} losses. In our work, we use supervised contrastive learning to build an character recognition model for ultra low scripts.

\section{Scripts and Datasets}
\subsection{The choice of script }
Our methodology is designed for scripts with very low digital resources, hence the choice of script to validate our methodology is an important decision. Takri script, has extremely low available digital resources, no labelled dataset along with low user base. George Grierson, in his Linguistic Survey of India, describes Takri and its variations as a script with shared inherent characteristics consequently classifying it as a "class of scripts" rather than a single script \cite{grierson1, grierson2}. 

To further validate the claim of having common linguistic characteristics within the dialects using Takri as a script, we use a set of 25 sentences used in day to day conversation, and translate them to seventeen dialects (used in the Himalayan regions of Himachal Pradesh, India) by the use of human annotators who are fluent in these dialects. We empirically study various semantic, lexical and syntactic features for these dialects and explore interdependence among these dialects using agglomerative hierarchical clustering \cite{agglomerative}. Appendix C illustrates the relationship between various dialects used in the Himalayan regions and how the use of one script binds all of them together. 

Takri, like most Indic script falls under \textit{Abugidas} class of writing systems \cite{writingSystems}, and some of the salient characteristics are summarised below:
\begin{itemize}
    \setlength\itemsep{-0.35em}
\item The character set of Takri comprises of 11 vowels, 33 consonants and 10 numbers.
\item There are 10 vowel modifiers which can occur on the top, below, left or right of the consonants. 
\item Takri script does not contain headline unlike other Indic scripts like Devanagari
  \item Half forms are not used in most versions of Takri. 
\item Ligatures are also infrequently written.  
\item Most characters consists of connected components only. 
\item Compound characters are not present in Takri.
 \end{itemize}

\subsection{Datasets}
To the best of our knowledge, there is only single source of a good quality dataset sourced from machine printed Takri books collected manually consisting of 272 text blocks containing 2,584 lines and 10,880 words with the resolution of 200dpi \cite{datasetTakri}. We use this dataset as the base and add more samples to increase volume and diversity of samples using data augmentation techniques.

For Gujarati, we use the machine printed limited dataset by Goswami et al. consisting of 7,221 symbols \cite{gujData}. For Modi, we use available dataset by Chandankhede et al \cite{modiData}. For Ol chiki and Wancho, there is no dataset available so we use the available printed books and build our own dataset. 

\section{VOLTAGE: The proposed methodology}
\textbf{V}ersatile contrastive learning based \textbf{O}CR methodology for ultra \textbf{L}ow-resource scripts \textbf{T}hrough \textbf{A}uto \textbf{G}lyph feature \textbf{E}xtraction (VOLTAGE) follows the pipeline of tasks including, pre-processing and segmentation, automated feature engineering and unsupervised labelling, data augmentation and classification, post processing and evaluation (see Figure \ref{Architecture}). We validate our results for Takri on end to end errors and character/word error rates. We further validate VOLTAGE for Modi, Ol Chiki, Gujarati and Wancho to establish the universal effectiveness of our work. We use Python 3.8.12 with conda, opencv for image processing along with ml libraries (keras, numpy, transformers) for our experiment.    

\begin{figure*}[t]
\centering
{
\fbox{\includegraphics[width=0.98\textwidth]{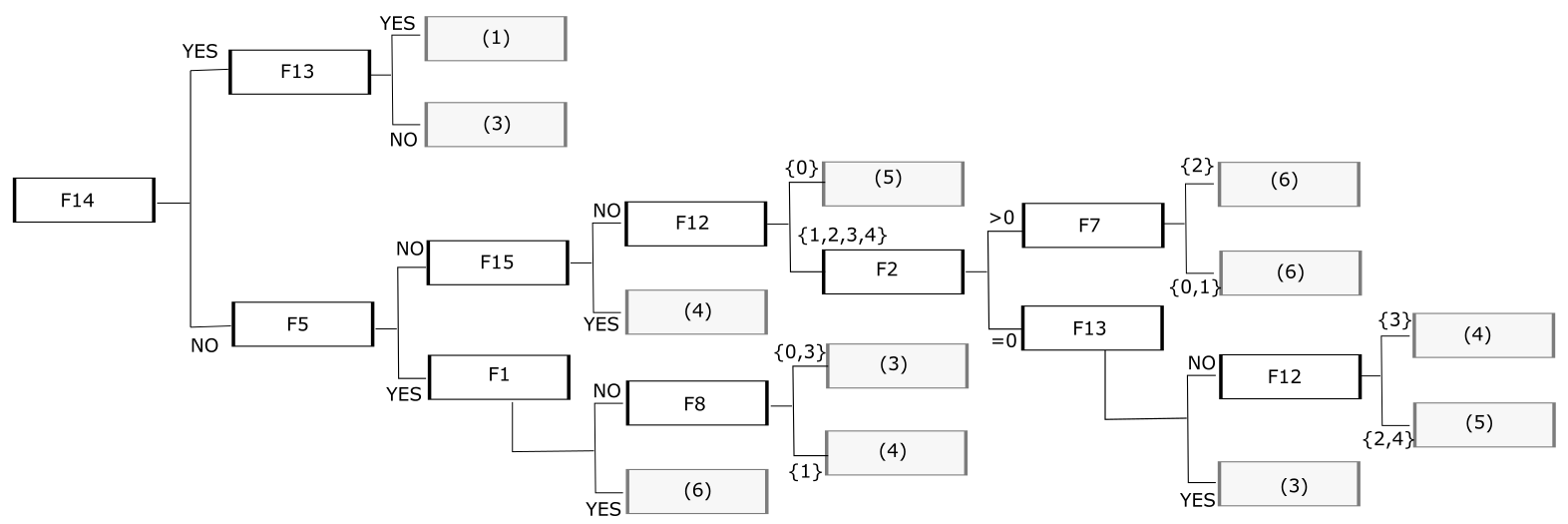}}
}
\setlength{\belowcaptionskip}{-12pt} 
\caption{Partition of characters into small groups based on glyph features as recommended by our design, where F$\#$ are feature identifiers in the glyph feature inventory}
\label{Clustering}
\end{figure*}

\subsection{Extraction}
It is imperative to extract and segment the input source into lines, words, characters and symbols before it can be put to use for downstream OCR tasks. Segmentation of page into lines and lines into words leverages computation of horizontal and vertical projections (HX and VY as illustrated in Eq. 1 and 2) and find valleys within the threshold \cite{linesegmentation, segmentation, classificationTakri}.

\begin{equation}
{VY_i}\:\: =  \sum_{x=0}^{x=Width} ({\text{No. of black pixels for $x_i$}})
\end{equation}
\begin{equation}
{HX_i}\:\: =  \sum_{y=0}^{y=Height} ({\text{No. of black pixels for $y_i$}})
\end{equation}

Segmentation of words into characters is slightly more complex due to close vicinity of characters and overlaps. To solve this issue, we have enhanced Eq. 2 and compute the enhanced horizontal projection (EHX) which applies additional penalty in downward direction for character segmentation (Eq. 3) because most overlaps in Takri occur in the upper parts. We have observed that this technique, helps in overall reduction of segmentation errors by 3\%. 

We have observed that abugidas class of scripts overlap their symbols (like Takri) and alphabetic scripts are isolated. Hence when we conduct our experiment for other scripts, we use EHX for Modi/Gujarati and HX for Ol chiki/Wancho.

\begin{equation*}
\begin{aligned}
{EHX_i}\:\: =  \sum_{y=0}^{y=Height} ({\text{No. of black pixels for $y_i$}}) 
\\
\vspace{-10pt}
+ ({\text{Penalty Wt. * $y_i$}}) \hspace{20pt} (3)
\end{aligned}
\end{equation*}

Furthermore, we further break the individual characters into sub-characters (also called as symbols) by dividing the space into three zones. We design a three step procedure to achieve this. (i) The first step is \textit{Skeletonization} which reduces the thickness of the character into single pixel, and helps to bring uniformity in the thickness irrespective of the input variation \cite{skeleton}; (ii) Once the characters obtain uniform thickness, we apply \textit{Connected Component Labelling} to label disjoint components \cite{connected}. Most symbols in Takri are connected (apart from few exceptions like \img{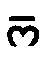} and hence this step helps in marking disconnected sections within the character. (iii) In the last step, we apply rule based method (to consider exceptions) and perform \textit{Zone Classification} and classify symbols in three different zones. This step is not needed for alphabetic scripts like English, and for those scripts characters and symbols are analogous to each other. The entire process of extraction is illustrated further in Appendix F.

\subsection{Annotation}

Takri consists of 50 symbols in middle zone, 6 in upper zone and 4 in bottom zone respectively. Annotation of extracted 14,000 symbols into appropriate category is a very critical activity for effective OCR design. Labelling each symbol manually may be the most accurate method but not scalable. Moreover, it becomes further tedious since there are only a handful of people who can read Takri. 

We perform unsupervised clustering of symbol images individually for each zone to label the dataset. Unsupervised clustering of images partitions the dataset into visually similar clusters without any access to ground truth labels. We use pretrained models on ImageNet and perform partition into individual characters \footnote{https://paperswithcode.com/task/image-clustering} \cite{scan}. We cluster and label images effectively with 96\% accuracy for upper and bottom zone characters (see Table \ref{tab:kmeans}). However the accuracy for middle zone characters was 69\% due to large spread of labels. It has been observed, that the errors have a linear dependency on the number of clusters \cite{kmeans}. This is also evident from our experiment, and hence it is advisable to divide middle-zone into smaller groups to overcome this.

\begin{table}[b]
  
  \caption{Unsupervised clustering accuracy for various zones for various k-means combinations.}
  \label{tab:kmeans}
\begin{adjustbox}{max width=.48\textwidth}
\begin{tabular}{l|ccc}
\toprule
{\textbf{}} &  \textbf {Upper} & \textbf {Middle} & \textbf {Bottom}\\
{\textbf{}} &  \textbf {Zone} & \textbf {Zone} & \textbf {Zone}\\
\hline
{Distribution} & {23\%} & {70\%} &  {7\%}\\
{No. of Labels} & {5} & {50} &  {4}\\
{Accuracy} &  &  & \\
\hspace{10pt} {50 iterations} & 91\% & 61\% & 93\%\\
\hspace{10pt} {300 iterations} & \textbf{96}\% & 69\% & \textbf{97}\%\\
{With feature recommendor} & - & \textbf{96\%} & -\\
 \hline
\bottomrule
\end{tabular}
\end{adjustbox}
\end{table}

\textbf{Glyph Feature Recommendation System (GFRS):} We design a novel recommender system "GFRS", which analyses and recommends the most appropriate glyph features for a given script from our inventory of glyph features without any human intervention. It takes a set of characters used in a language as input and recommends a tree structure with the recommended glyph features, and distribute the characters into smaller groups for more effective annotation. We have validated this for Takri along with 4 other Indic scripts. As illustrated in Figure \ref{Clustering}, when GFRS is applied on Takri it recommends 9 features (F1: Presence of headline; F2: Number of loops; F5: Presence of right sidebar; F7: Number of endpoints; F8: Number of junctions; F12: Aspect ratio; F13: Horizontal symmetry; F14: Vertical symmetry; and F15: Number of dots) from the feature store and each identified subgroup does not contain more than 6 symbols. Our process of building the feature store is iterative after analysing the shape characteristics of multiple Indic scripts. We have observed that using the approach, the unsupervised labelling accuracy improves to 96\% for middle zone (which was 69\% earlier).

Appendix A illustrates the entire list of glyph features (32 in total count), which forms the inventory of shape feature set (analysing the distinctive characteristics of these glyphs like number of loops, lines, endpoints, junctions, symmetry etc.). GRFS recommends the most appropriate feature set for a particular script which helps in appropriate distribution of characters and facilitate automatic labelling. We also illustrate as part of Appendix A, the various recommended feature sets for other scripts used in our paper including Modi, Ol Chiki, Gujarati and Wancho.

\begin{table}[b]
  \caption{High level summary on multiple error metrics for Takri dataset.}
  \label{tab:ResultaSummary}
\begin{tabular}{p{4.5cm}|p{2cm}}
\toprule
{\textbf{Error Metric}} & \textbf {VOLTAGE} \\
 \hline
 {E2E (End to End)} & {18\%} \\ 

 {WER (Word Error)} & {12\%} \\ 

 {CER (Character Error)} & {4\%} \\ 
\bottomrule
\end{tabular}
\end{table}

\subsection{Re-enforcement}
\begin{figure}%
    \centering
    \subfloat {\includegraphics[width=7.5cm]{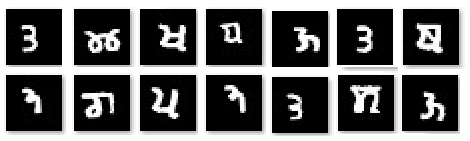} }%
    
    \subfloat{{\includegraphics[width=7.5cm]{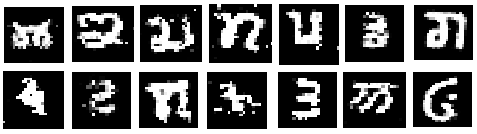} }}%
   \vspace{-6pt}
   \setlength{\belowcaptionskip}{-15pt} 
    \caption{Samples from augmented characters (row 1 and 2) using image transformations (row 3 and 4) using GAN}%
    \label{fig:example}%
\end{figure}

Our labelled dataset contain approximately 14,000 symbols. We augment this dataset applying transformations for rotations, sheer and brightness. We limit the angular transformation to $9^o$, and 10\% range on sheer and brightness. Further, we use the transformed images and build a GAN for each character, by the use of four layer generator and discriminator networks, with a learning rate of $2$ x $10^{-4}$ and train on 400 epochs. Figure \ref{fig:example} illustrates the examples from this. The final dataset contains 225,000 symbols for Takri.

\subsection{Identification}

The earlier forms of OCR designs, either use CNN \cite{cnn} or a combination of CNN and LSTM \cite{lstm} as a deep learning method for character identification. We use supervised contrastive learning for our classifier, leveraging multiple positive and negative samples. We discuss the benefits of using this approach empirically later in this paper. Table \ref{tab:ResultaSummary} illustrates multiple error metrics in our work. 

Appendix E illustrates details on the architecture of contrasting learning used in our work. We  illustrate how transformations (including image processing and GAN) are put to use in an encoder model which maps them to a latent representation space, encapsulating features and similarities. We apply supervised contrastive loss function from SupCon, to maximise the agreement between positive pairs (same character images) and minimize the agreement between negative pairs (different character images) \cite{SupCon}. 
\begin{equation*}
\begin{aligned}
{L}\:\: =  \sum_{i\in I } \frac {-1}{\left|P(i)\right|} \sum_{p \in P(i)} log \frac{exp (z_i.z_p/\tau)}{\sum_{a \in A(i)} exp(z_i.z_a/\tau)} 
\end{aligned}
\end{equation*}

Here $z_l = Proj(Enc(\tilde{x}_l)) \in R^{Dp}$, the $\cdot$ symbol denotes inner dot product, $ \tau  \in R^+$ is scaler temperature parameter. Index $i$ is called anchor, index $j(i)$ is called positive and other indices are called negative.   $P(i) \equiv { p \in A (i): \tilde{y_p} = \tilde{y_i}}$ is the set of indices of all positive in multiviewed batch (2N augmented samples) distinct from $i$, and $|P(i)|$ is its cardinality.

\begin{table}[t]
  \caption{Misinterpretation of characters, due to similarity of glyph or part and whole relationships}
  \label{tab:Misinterpretation}
\begin{adjustbox}{max width=.48\textwidth}
\renewcommand{\arraystretch}{0}
\begin{tabular}{p{1.2 cm}| p{1.9 cm} | p{2.9 cm}}
\toprule
\textbf{Actual} & \textbf {Recognised} & \textbf {Type of error}\\
 \hline
  {\includegraphics[trim=0 0 0 -5 width = 0.8 cm, height = 0.6 cm]{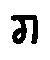}} &
  {\includegraphics[trim=0 0 0 -5 width = 0.8 cm, height = 0.6 cm]{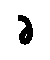}} {\includegraphics[trim=0 0 0 -5 width = 0.8 cm, height = 0.6 cm]{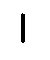}} &
  \\

  {\includegraphics[trim=0 0 0 -5 width = 0.8 cm, height = 0.6 cm]{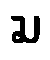}} &
  {\includegraphics[trim=0 0 0 -5 width = 0.7 cm, height = 0.6 cm]{img/RA.png}} {\includegraphics[trim=0 0 0 -5 width = 0.8 cm, height = 0.6 cm]{img/VIRAM.png}} & {Over Segmentation} 
   \\

 {\includegraphics[trim=0 0 0 -5 width = 0.8 cm, height = 0.6 cm]{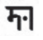}} &
  {\includegraphics[trim=0 0 0 -5 width = 0.8 cm, height = 0.65 cm]{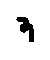}} {\includegraphics[trim=0 0 0 -5 width = 0.8 cm, height = 0.6 cm]{img/VIRAM.png}} &
   \\
   \\
   \hline
  {\includegraphics[trim=0 0 0 -5 width = 0.8 cm, height = 0.6 cm]{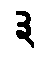}} &
  {\includegraphics[trim=0 0 0 -5 width = 0.8 cm, height = 0.65 cm]{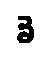}} &
   \\
  
    {\includegraphics[trim=0 0 0 -5 width = 0.8 cm, height = 0.6 cm]{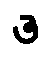}} &
  {\includegraphics[trim=0 0 0 -5 width = 0.8 cm, height = 0.6 cm]{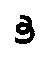}} & {{Mis Classification} }   
   \\
   
      {\includegraphics[trim=0 0 0 -5 width = 0.65 cm, height = 0.5 cm]{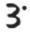}} &
  {\includegraphics[trim=0 0 0 -10 width = 0.9 cm, height = 0.65 cm]{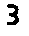}} &
   \\   
   \hline
\bottomrule
\end{tabular}
\end{adjustbox}
\end{table}

\subsection{Post Processing}
Table \ref{tab:Misinterpretation} illustrates some examples where the characters are misinterpreted due to over segmentation of characters or incorrect classification due to similarity in visual characteristics. It is therefore pivotal to have some post processing and correct these errors based on language grammar and patterns. Appendix B illustrates  an inventory of principles and guidelines for Indic languages. These principles help in improving the overall accuracy of recognised text, considering linguistic context along with the syntax. We have identified a set of generic and specialised linguistic rules for Indic scripts, and apply them towards the end of our pipeline.
\section{Results and Discussion}

Quantification of errors in OCR pipeline, specially for very low resource scripts is ambiguous unless properly defined \cite{err2}. The document page needs to be segmented into lines, words, characters and symbols for OCR engine. Any error which occurs in this step would fall under layout segmentation error. Most OCR literature do not include segmentation errors as part of OCR errors. 

OCR systems generally consider errors either for the entire End to End (E2E) pipeline, at word level or character levels. E2E includes errors at various stages of pipeline such as Pre-processing, Segmentation, Classification, and Post processing. Word recognition accuracy or Word Error Rate (WER) is the average percentage of mis-recognised words. Character Errors Rate (CER) is the ratio of mis-recognised symbols within accurately segmented symbols. 

\begin{equation*}
\text{WER} = \frac{N_w'}{N_w}, \quad
\text{CER} = \frac{N_s'}{N_s}
\end{equation*}

\noindent
where $N_w'$ is count of mis-recognised words, $N_w$ is correctly segmented words, $N_s'$ is count of mis-recognised symbols, and $N_s$ is correctly segmented symbols.

The concept for character/symbol is also ambiguous unless defined clearly, due to the linguistic peculiarity for each script. Symbols are monolithic for \textit{Alphabetic} languages (like English) which uses each symbol in same form, however for \textit{Abugida} scripts (like Indic scripts) symbols are of multiple types, namely (a) Root Symbols: like \img{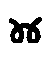} ("Ka") and \img{img/GA.png} ("Ga") and (b) Modifier/ Marker Symbols: like vowel modifiers \img{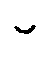} ("U") and \img{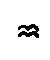} ("Au") \cite{writingSystems}. We consider root and markers separately independent of each other in our empirical study. An error in root symbol does not contribute to the error in associated marker symbols.

\begin{table}[t]
  \caption{Empirical study for VOLTAGE on Takri on Machine Printed (MP) and Hand Written (HW) samples.}
  \label{tab:ResultTakri}
\begin{tabular}{p{3.0 cm} | p{1.5 cm} p{1.5 cm}}
\toprule
{\textbf{Zone}} & \textbf {MP} & \textbf {HW} \\
 \hline
 {UZ (Upper Zone)} & {96\%} & {88\%} \\ 
 {MZ (Middle Zone)} & {94\%} & {85\%}\\ 
 {BZ (Bottom Zone)} & {97\%} & {89\%}\\ 
\bottomrule
\end{tabular}
\end{table}

\begin{table*}
  \caption{Evaluation across other scripts. For Gujarati we experimented with two scenarios, (a) Gujarati LRL- Like low resource language and (b) Gujarati HRL- like high resource language}
  \label{tab:otherOCR}
\begin{tabular}{p{2.3 cm} p{1.5 cm} p{1.5 cm} p{1.6 cm} p{1.6 cm} p{1.5 cm} p{1.3 cm} p{1.4 cm}}
\toprule
\multirow{ 2}{*}{\textbf{{Script Name}}} & \multirow{ 2}{*}{\textbf {{Script type}}} & \multirow{ 2}{*}{\textbf {{Language}}} & \textbf {{SOTA}} & \textbf {{VOLTAGE}} & \textbf {{Dataset}} & \textbf {{Label}} & \textbf {{Glyph}} \\
 &  &  & \textbf {{Accuracy}} & \textbf {{Accuracy}} & \textbf {{size}} & \textbf {{count}} & \textbf {{features}} \\
\hline
Takri & Abugida & Multiple & NA & 95\% & 14,051 & 59 & 9\\
Modi & Abugida & Marathi & (84-94)\% & 93\% & 7,221 & 46 & 11\\
Ol Chiki & Alphabet & Santali & (83-92)\% & 91\% & 8,873 & 30 & 5\\
Gujarati LRL & Abugida & Gujarati & (86-96)\% & 93\% & 7,643 & 42 & 9\\
Gujarati HRL & Abugida & Gujarati & (86-96)\% & 96\% & 200K+ & 42 & 9\\
Wancho & Alphabet & Wancho & NA & 91\% & 6,500 & 42 & 8\\
\bottomrule
\end{tabular}
\end{table*}

\begin{figure}[b]
\centering
{
\fbox{\includegraphics[width=0.45\textwidth]{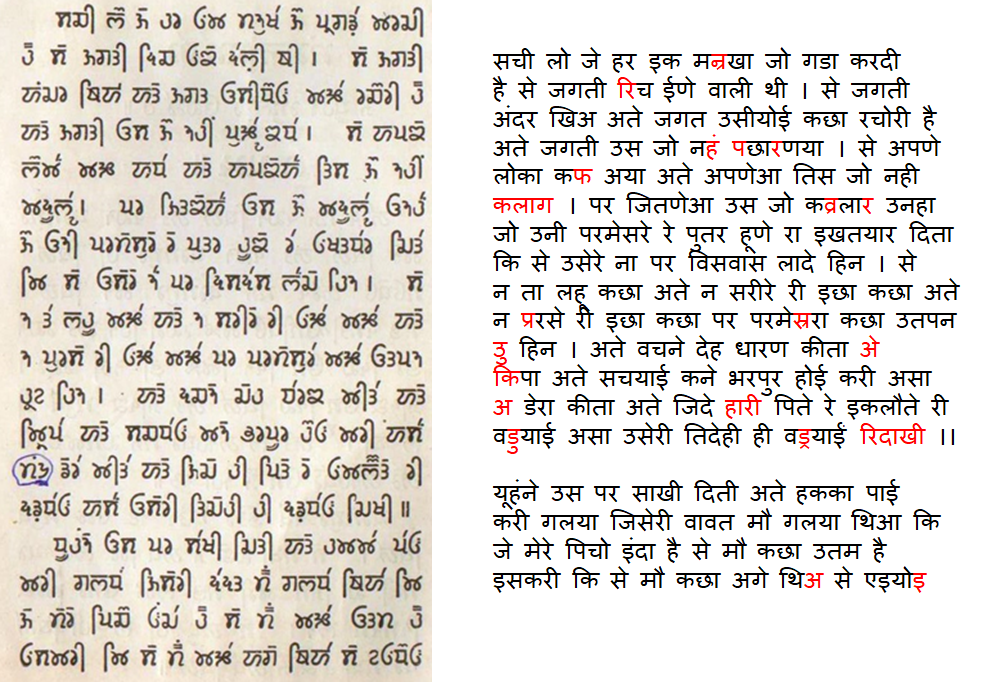}}
}
\caption{Applying OCR at page level.}
\label{Final}
\end{figure}

Most OCR studies, considers errors within (0-2\%), (2-10\%) and (>10\%) as good, average and Poor \cite{cer}, respectively. However for Indic scripts with limited training data and unknown vocabulary along with heterogeneous handwritten forms, a CER value as high as around (10-20)\% is considered satisfactory \cite{indicocraccuracy, handocr}. We compute many error metrics (see Table \ref{tab:ResultaSummary}) but observed that all metrics are co-related. With this observation and existing practices, we use the standard metric, CER, for further evaluation not including errors during segmentation and pre-processing \cite{cer}.

\subsection {Empirical study on Takri}

We evaluate our results both for \textit{machine printed} and \textit{handwritten} samples. We identify a group of 22 participants (13 male, 9 female; diverse age groups; belonging to Himachal Pradesh) who were made familiar with our work and Takri characters. We asked them to record symbols and label them.  

We also evaluate our results separately for each zone and analyse the results. As illustrated in Table \ref{tab:ResultTakri} we make the following observations, (a) The upper-zone symbols which account for approx. 16\% of corpus have recognition accuracy of 96\% (88\% for handwritten samples). (b) The middle-zone symbols account for the majority of characters and is most busiest zone. This contributes for approx 79\% of symbols and recognition accuracy is 94\% (85\% for handwritten samples). (c) The bottom-zone is the most infrequent zone with approx. 5\% symbols and recognition accuracy is 97\% (89\% for handwritten samples).

 Figure \ref{Final} presents an illustrative example containing 18 lines with 162 words and we pass this via VOLTAGE. We observe that for this sample, 19 words are mis-recognized in recognised output (illustrated in red in the figure), thereby getting E2E error of 12\%. 

\subsection {Applying the methodology to other scripts}

We apply our methodology to four other diverse Indic scripts (\textit{Modi, Ol Chiki, Gujarati and Wancho}) to validate the overarching effectiveness of our work. While Modi and Gujarati belong to abugidas family of scripts with ancient history, Ol Chiki and Wancho are more modern alphabetic scripts. Modi, Ol Chiki and Wancho are very low on resources and less explored \cite{OlChikiOCR, ModiOCR, Wancho}. Gujarati is more popular and resourceful but we use only limited data for our experiment, treating this as low resource experiment \cite{gujShape, gujStroke} and also as HRL. 

The purpose of using a mix of Indic scripts is to validate the all-inclusive application of our design across multiple types of scripts.  It is evident from our results in Table \ref{tab:otherOCR} that VOLTAGE generalizes well as it provides consistent results across scripts and can be very useful for scripts where labelled data is a scarce. 

\begin{table}[b]
\caption{Applying annotated Takri dataset to other script OCR, and evaluating accuracy.}
\label{tab:AbTransfer}
\begin{tabular}{p{2.1 cm} | p{0.8 cm} p{0.8 cm} p{0.8 cm} p{0.8 cm}}
\toprule
{\textbf{Foundation}} & \multicolumn{4}{c}{\textbf{Training samples (Takri)}}\\
{\textbf{Script}} &  \textbf {10K} & \textbf {30K} & \textbf {50K} & \textbf {100K}\\
 \hline
 {Devanagiri} & {59\%} & {81\%} & {83\%} & {86\%}\\ 
 {Gurumukhi} & {54\%} & {79\%} & {81\%} & {86\%}\\ 
 {Gujarati} & {61\%} & {83\%} & {84\%} & {87\%}\\ 
 {Oriya} & {56\%} & {79\%} & {82\%} & {87\%}\\ 
 {Bangali} &  {58\%} & {74\%} & {81\%} & {85\%}\\ 
 {Tamil} & {39\%} & {61\%} & {71\%} & {78\%}\\ 
\bottomrule
\end{tabular}
\end{table}

\subsection {Baseline studies}

\begin{table*}
  \centering
  \caption{ Symbol counts in Takri dataset and models used}
  \label{tab:Dataset}
\begin{tabular}{c|l|cc|cc|cc}
\toprule
\multirow {2}{*}{\textbf{Sr. No.}} & \multirow {2}{*}{\textbf{Transformations}} & \multicolumn{2}{c}{\textbf {UZ}} & \multicolumn{2}{c}{\textbf {MZ}} & \multicolumn{2}{c}{\textbf {BZ}}\\
& & \textbf {count}  & \textbf {model} & \textbf {count}  & \textbf {model} & \textbf {count}  & \textbf {model}\\
 \hline
 1 & None (Actual images) & 2981 & $B^{uz}$  & 9702 & $B^{mz}$ & 1368 & $B^{bz}$\\

 2 & Image Transformations & 28,563 & $E_1^{uz}$ & 110,012  & $E_1^{mz}$ & 15,931 & $E_1^{bz}$\\ 

 3 & Generative Images  & 40,427 & $E_2^{uz}$ & 164,162 & $E_2^{mz}$ & 21,403  & $E_2^{bz}$\\ 
\hline
\end{tabular}
\end{table*}

We use the available annotated data-sets for multiple high resource Indic script and fine tune for Takri \cite{akshara}. As illustrated in Table \ref{tab:AbTransfer} we observe the following issues with this approach, (a) The choice of what foundation script to choose is very important. In case of Takri we observed using Gujarati gives best results. The choice of foundation model has lot of manual intervention hence we did not include this in the overall process. (b) We restricted our few shot experiment till 100K, since moving to higher numbers would need lot of data which is not feasible for ultra low scripts, also leads to catastrophic interference thereby defying the purpose of using a foundation model. We also see that as the number of samples go up, most models converge to similar results, and far from that of obtained by VOLTAGE. Appendix G further illustrates the size and source of individual data sets across multiple scripts. We also illustrate the SOTA accuracy for these data rich scripts and compare with the results we have achieved.

\subsection {Ablation studies}
We conduct ablation studies for VOLTAGE to substantiate the contributions of individual elements along with improved model understanding. We already illustrated in Section 4.2 that the application of GFRS improves annotation accuracy by 27\%. We conduct more experiments to substantiate the importance of each step in the whole pipeline.

We use basic CNN-LSTM models on each zone separately and test them. We train three separate sets of models for each zone. Within each zone there are separate models subject to the source of training data used, thereby resulting in total of nine models (see Table \ref{tab:Dataset}). These models can be classified as (i) three Base models one for each zone ($B^{uz}$, $B^{mz}$, $B^{bz}$) which include only the actual images extracted from source documents for training, (ii) three Enriched$_1$ models ($E_1^{uz}$, $E_1^{mz}$, $E_1^{bz}$) which include data generated by image transformations along with data used in base models, and (iii) three Enriched$_2$ models ($E_2^{uz}$, $E_2^{mz}$, $E_2^{bz}$) which include images generated by GAN along with the data used in $E_1$ models. We compare these nine models with three models (one for each zone) using contrastive learning as described in our work ($V$).

As illustrated in Table \ref{tab:Results1} we observe that VOLTAGE models outperforms the base models for both machine printed and handwritten evaluation.

\begin{table}[h]
  \caption{Character error rates (CER) for Machine Printed (MP) and Handwritten (HW) symbols for CNN-LSTM models and compare the results with VOLTAGE models (V)}
  \label{tab:Results1}
\begin{adjustbox}{max width=.48\textwidth}
\begin{tabular}{p{2.0 cm} | p{0.9 cm} p{0.9 cm} p{0.9 cm} p{0.9 cm}}
\toprule
\textbf & \textbf{Model} & \textbf {{UZ}} & \textbf {{MZ}} & \textbf {{BZ}}\\
\hline
{{Composition}} & &  {16}\%  & {79\%}  & {5}\%\\
\hline
 & B &  {06\% }  & {08\%}  & {06\%}\\
\multirow{ 2}{*}{CER-MP} & {$E_1$} &   {04\%}  & {07\% }  & {03\% }\\
 & $E_2$ &  {08\% }  & {09\% }  & {10\% }\\
 & $V$ &  {04\% }  & {06\% }  & {03\% }\\
\hline 
 & B &  {21\%  }  & {27\% }  & {21\% }\\
\multirow{ 2}{*}{CER-HW} & $E_1$ & {19\% }  & {25\% }  & {19\% }\\
 & {$E_2$}&  {14\% }  & {17\% }  & {15\% }\\
 & $V$ &  {12\% }  & {15\% }  & {11\% }\\
\bottomrule
\end{tabular}
\end{adjustbox}
\end{table}

\subsection {Use Cases: Takri for the digital world}
We have observed that there is dearth of printed books on Takri, hence it is important to facilitate Takri in printed form. We facilitate NLP for Takri by developing two ready to use tools, (a) Transliteration to Takri facilitating digitization of folk literature (via development of standardised individual symbol images, and creating rule based engine to amalgamation) and (b) Synthetic generative models for each symbol in Takri.  

Appendix D further illustrates with an example, how our transliteration engine converts text in other languages to Takri in digital format. This can be very instrumental to publish small stories, news headlines etc. and shared in interested community to facilitate the use of the script. We also share our GAN models to be used by fellow researchers for furthering research\footnote{https://github.com/prawaal/Takri}.

\section{Conclusion and Future directions}
The paper presents a comprehensive unsupervised OCR methodology, VOLTAGE, which includes a novel Glyph feature recommendation system (GFRS) for effective symbol labelling. We developed VOLTAGE using a very low resource language, Takri, and validated its effectiveness and generalization on various Indian scripts. We achieve accuracy at par with SOTA of respective test scripts. We also build use cases for Takri to demonstrate the usefulness of the work. Our work can facilitate the digitization of ultra low resource scripts thereby save them from extension.

As part of our future work, we shall use our method to build more comprehensive datasets along with building the vocabulary for Indic languages to help in error correction during post processing. We also plan to use the method as described to digitize more languages within India partnering with local governments.

\section*{Limitations}
Glyph feature store is designed keeping in mind, the stroke characteristics of Indic scripts. Hence it can work for all Indic scripts without any modifications, but may need changes for other family of scripts. It is possible to use same design principles and extend the feature stores for any other family of scripts and apply the method as described in our paper.

\bibliography{custom}
\bibliographystyle{acl_natbib}

\section*{Acknowledgements}
We would like to thank Shikha Magotra for compiling the digital database of \textit{Takri} script from multiple museums in north-western Indian geography and making it available for our research. 

\newpage
\appendix

\onecolumn

\label{appendixA}
\section*{Appendix A: Glyph feature store}
\begin{longtable} [p]{| p{.1\textwidth} | p{.65\textwidth} | p{.07\textwidth} | p{.07\textwidth}|} 
 \hline
\textbf{Feature ID} & \textbf{Feature Description} & \textbf{Type} & \textbf{Range} \\[0.5ex]
\hline
F1 & \textbf{Presence of Headline}: Checks for the presence of a horizontal line on the top of the sub-symbol. & Boolean & 0/1\\
\hline
F2 & \textbf{Number of Loops}: Counts the number of loops in the symbol including loops with headline. & Number & 0-N\\
\hline
F3 & \textbf{Number of Loops with headline}: Counts the number of loops the symbol makes with the headline (F1). & Number & 0-N\\
\hline
F4 & \textbf{Presence of left-sidebar}: Check for the presence of a vertical line on the left-most side of sub-symbol. & Boolean & 0/1\\
\hline
F5 & \textbf{Presence of right-sidebar}: Check for the presence of a vertical line on the right-most side of sub-symbol. & Boolean & 0/1\\
\hline
F6 & \textbf{Number of connected components}: Counts the number of sub-symbols which are connected. & Number & 0-N\\
\hline
F7 & \textbf{Number of endpoints}: Counts the number of points, which have only one black pixel in its 3x3 neighbourhood. & Number & 0-N\\
\hline
F8 & \textbf{Number of junctions}: Counts the number of points, which have more than two black pixel in its 3x3 neighbourhood. & Number & 0-N\\
\hline
F9 & \textbf{Number of junctions with headline}: Counts the number of junctions (F8) which touch the headline (F1). & Number & 0-N\\
\hline
F10 & \textbf{Number of bend points clockwise}: Counts the number of points which makes 90 degree turn towards right direction. & Number & 0-N\\
\hline
F11 & \textbf{Number of bend points anti-clockwise}: Counts the number of points which makes 90 degree turn towards left direction. & Number & 0-N\\
\hline
F12 & \textbf{Aspect Ratio}: Ratio of symbol height and width on 0-100 scale. & Number & 0-100\\
\hline
F13 & \textbf{Horizontal Symmetry}: Flip the image on Y axis (mirror the image in left-right perspective). Find the similarity with original image using threshold value. & Boolean & 0/1\\
\hline
F14 & \textbf{Vertical Symmetry}: Flip the image on X axis (mirror the image in top-down perspective). Find the similarity with original image using threshold value. & Boolean & 0/1\\
\hline
F15 & \textbf{Number of Dots}: Counts the number of points, which have zero black pixels in its 3x3 neighbourhood. & Number & 0-N\\
\hline
F16 & \textbf{Number of left-right layers}: Counts the number of layers of black pixels on x axis. This relates to the maximum isolated black pixels in horizontal cross section. & Number & 0-N\\
\hline
F17 & \textbf{Number of top-down layers}: Counts the number of layers of black pixels on y axis. This relates to the maximum isolated black pixels in vertical cross section. & Number & 0-N\\
\hline
F18 & \textbf{Minimum horizontal projection}: Compute the horizontal projection (count the number of black pixels across y axis for every x axis) and find the minimum value. Scale this by taking ratio with height of image, and multiple by 100, & Number & 0-100\\
\hline
F19 & \textbf{Minimum vertical projection}: Compute the vertical projection (count the number of black pixels across x axis for every y axis) and find the minimum value. Scale this by taking ratio with width of image, and multiple by 100, & Number & 0-100\\
\hline
F20 & \textbf{Maximum horizontal projection}: Compute the horizontal projection (count the number of black pixels across y axis for every x axis) and find the maximum value. Scale this by taking ratio with height of image, and multiple by 100, & Number & 0-100\\
\hline
F21 & \textbf{Maximum vertical projection}: Compute the vertical projection (count the number of black pixels across x axis for every y axis) and find the maximum value. Scale this by taking ratio with width of image, and multiple by 100, & Number & 0-100\\
\hline
F22 & \textbf{Maximum left depth}: The maximum depth of the left profile calculated
as a percentage with respect to total width of the box enclosing the symbol.  & Number & 0-100\\
\hline
F23 & \textbf{Maximum right depth}: The maximum depth of the right profile calculated
as a percentage with respect to total width of the box enclosing the symbol.  & Number & 0-100\\
\hline
F24 & \textbf{Maximum top depth}: The maximum depth of the top profile calculated
as a percentage with respect to total width of the box enclosing the symbol.  & Number & 0-100\\
\hline
F25 & \textbf{Maximum bottom depth}: The maximum depth of the bottom profile calculated
as a percentage with respect to total width of the box enclosing the symbol.  & Number & 0-100\\
\hline
F26 & \textbf{Minimum left depth}: The minimum depth of the left profile calculated
as a percentage with respect to total width of the box enclosing the symbol.  & Number & 0-100\\
\hline
F27 & \textbf{Minimum right depth}: The minimum depth of the right profile calculated
as a percentage with respect to total width of the box enclosing the symbol.  & Number & 0-100\\
\hline
F28 & \textbf{Minimum top depth}: The minimum depth of the top profile calculated
as a percentage with respect to total width of the box enclosing the symbol.  & Number & 0-100\\
\hline
F29 & \textbf{Minimum bottom depth}: The minimum depth of the bottom profile calculated
as a percentage with respect to total width of the box enclosing the symbol.  & Number & 0-100\\
\hline
F30 & \textbf{Stroke length}: Count of total black pixels as a percentage with total area (height * width) of the symbol box.  & Number & 0-100\\
\hline
F31 & \textbf{Epicenter top down}: Find the mean of all black pixels, and compute the location of Y axis from center as a percentage from mid point on y axis .  & Number & 0-100\\
\hline
F32 & \textbf{Epicenter left right}: Find the mean of all black pixels, and compute the location of X axis from center as a percentage from mid point on x axis .  & Number & 0-100\\
\hline
\end{longtable}
\textbf{Recommended Feature set for languages:}
\begin{itemize}
    \setlength\itemsep{-0.35em}
\item Takri - F1, F2, F5, F7, F8, F12, F13, F14, F15.
\item Modi - F1, F2, F4, F5, F7, F9, F12, F15, F16, F23, F30 .
\item Ol Chiki - F2, F4, F8, F12, F16.
\item Gujarati - F1, F2, F4, F5, F7, F12, F13, F15, F16.
\item Wancho - F2, F5, F8, F12, F13, F15, F16, F21.
\end{itemize}

\newpage

\section*{Appendix B: Post processing rule inventory} 
In our work for Takri OCR, we have applied rules from R1 to R7. We could not apply R8, R9, R10 due to non availability of dictionary for Takri.
\begin{longtable} [p]{| p{.1\textwidth} | p{.45\textwidth} | p{.34\textwidth}|} 
 \hline
 \textbf{Rule ID} & \textbf{Rule description} & \textbf{Remarks} \\[0.5ex]
\hline
R1 & One consonant should not contain more than one vowel modifier. &  \\
\cline{1-2}
R2 & Sentence delimiter should not happen in between of a word. &  \\
\cline{1-2}
R3 & Sentence delimiters should not repeat consecutively. & \\
\cline{1-2}
R4 & Numbers should not merge with consonants or vowels within the word boundary. & \\
R5 & Independent vowels symbols (excluding the modifiers) can occur only at the start of end of the word in most Indic scripts. &  Generic rule for Indic scripts. However some fine tuning may be needed from script to script basis.\\
\cline{1-2}
R6 & \textit{Panchamkshar} Characters (like  \img{img/NYA.png} ("Nya") and \img{img/NGA.png} ("Nga") in case of Takri) should not occur in the end of the word. &  \\
\cline{1-2}
R7 & Unicode sequence of symbols identified may not follow the same sequence as required by computing systems, so sequence of unicodes may need to be changed. For example \img{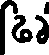} ("Fira") identifies vowel "i" before consonant "f" and sequence needs to be changed. &  \\
\hline
R8 & Shape characteristics of some symbols (glyph features) may follow Whole-Part design. Aggregation of some symbols (parts) may together form another symbol (whole). For example in Takri the "parts" - \img{img/RA.png} ("Ra") and \img{img/VIRAM.png} ("Viram") can form "whole" - \img{img/GA.png} ("Ga"). &  \\
\cline{1-2}
R9 & Shape characteristics of some symbols may be very similar and create mis-classification. For example \img{img/NGA.png} ("Nga") and \img{img/THREE.png} ("Number 3") are very similar  & {Validation with prebuilt vocabulary/dictionary to generate/shortlist candidate word.}\\
\cline{1-2}
R10 & It is possible that in case of short word partitions, the boundaries of word are not marked correctly. In case the word spans across two lines (with not sufficient space at the end of the line) word partition error can happen as well.  & \\
\hline
\end{longtable}

\appendix

\newpage

\label{appendixC}
\section*{Appendix C: Languages using Takri as a script and their relationships}
\begin{figure*}[h]
\centering
{
\fbox{\includegraphics[width=0.95\textwidth]{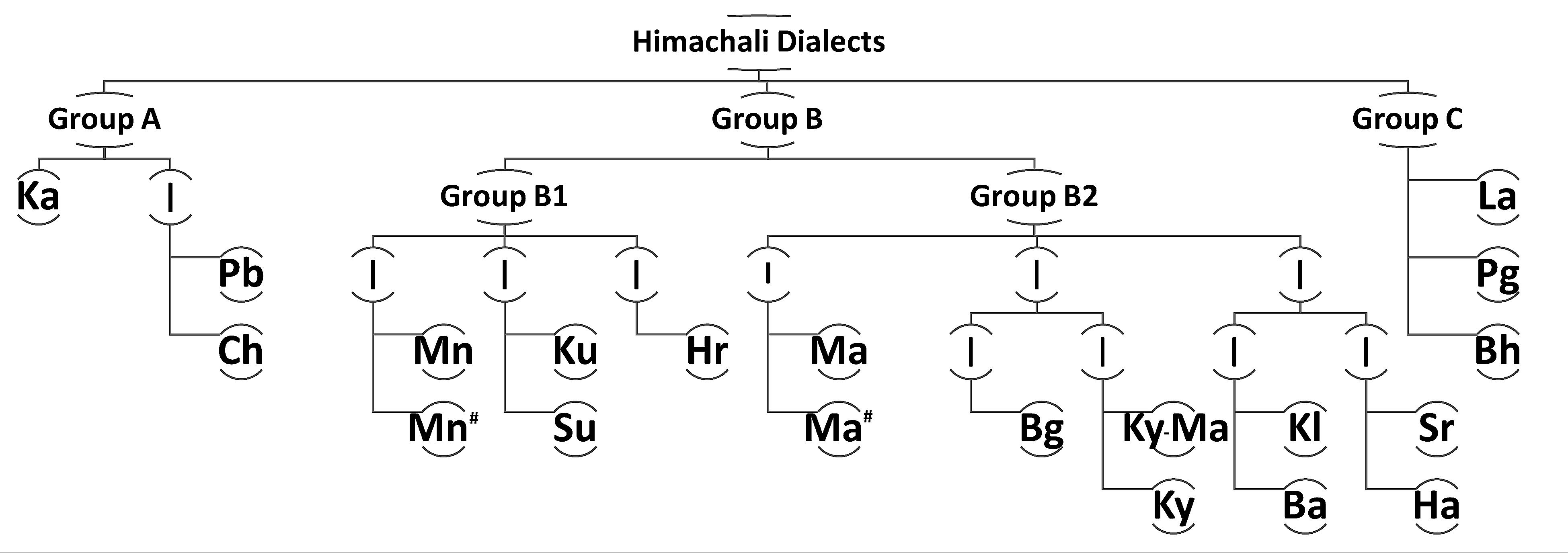}}
}
\end{figure*}
\noindent
Dialects with inflections are represented with superscript ('). (Baagli (Ba), Bhagati (Bg), Bharmauri (Bh), Chambiali (Ch), Hamirpuri (Hr), Hatti (Ha), Kalhuri (Kl), Kangri (Ka), Kinnauri (Kn), Kulluvi (Ku), Kyunthali (Ky),  Lahauli (La), Mahasuvi (Ma), Mandiali (Mn), Pangwali (Pg), Punjabi (Pb),  Sirmauri (Sr), Suketi (Su))

\label{appendixD}
\section*{Appendix D: Transliteration Sample to Takri using custm made glyhs.}
\begin{figure*}[h]
\centering
{
\fbox{\includegraphics[width=0.95\textwidth]{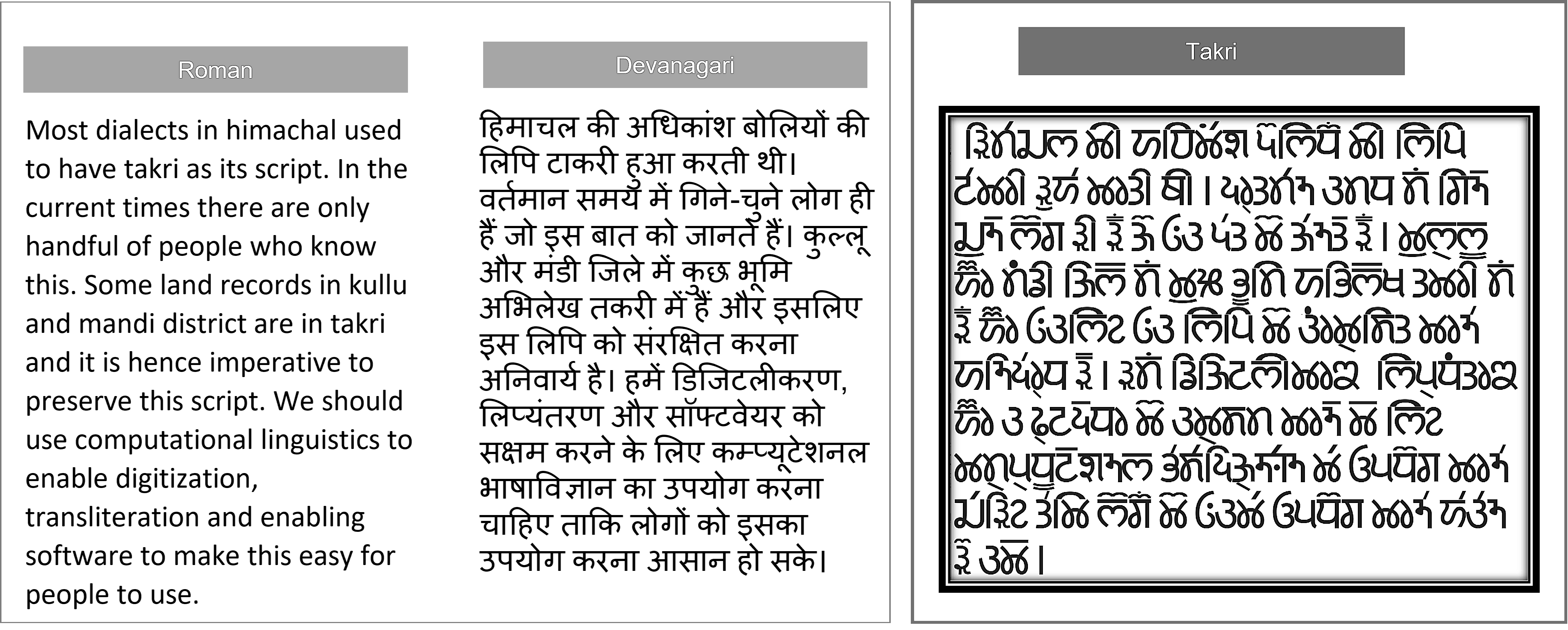}}
}
\end{figure*}

\newpage
\label{appendixE}
\section*{Appendix E: Use of supervised contrastive learning for character identification}
\begin{figure*}[h]
\centering
{
\fbox{\includegraphics[width=0.95\textwidth]{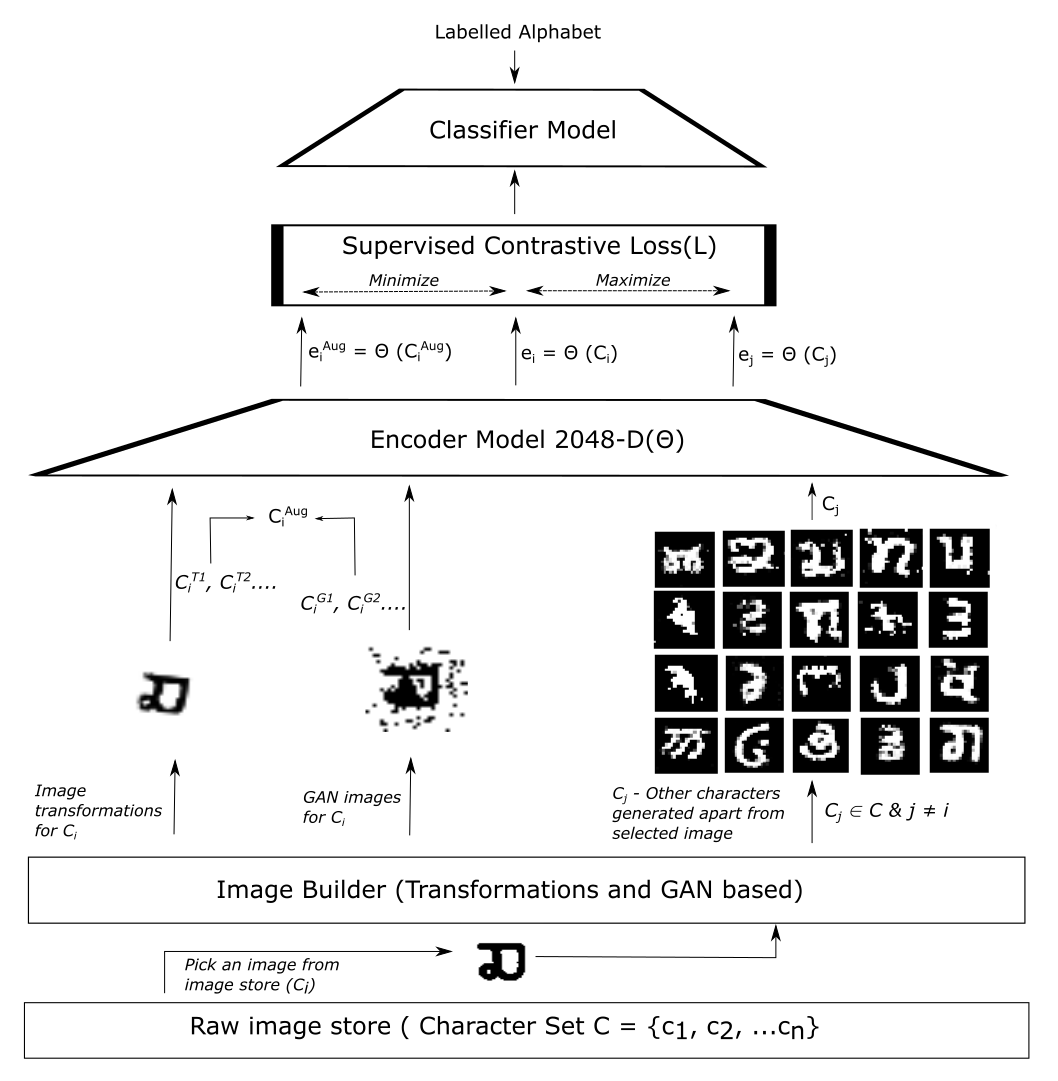}}
}
\end{figure*}
We use Google colab for running our experiment. For GAN images we train for 150 epochs, 4 layer generator and network, adam optimiser and learning rate of 0.0002.

\newpage
\label{appendixF}
\section*{Appendix F: Extraction of source raw data into lines, words, characters and symbols.}
\begin{figure*}[h]
\centering
{
\fbox{\includegraphics[width=0.95\textwidth]{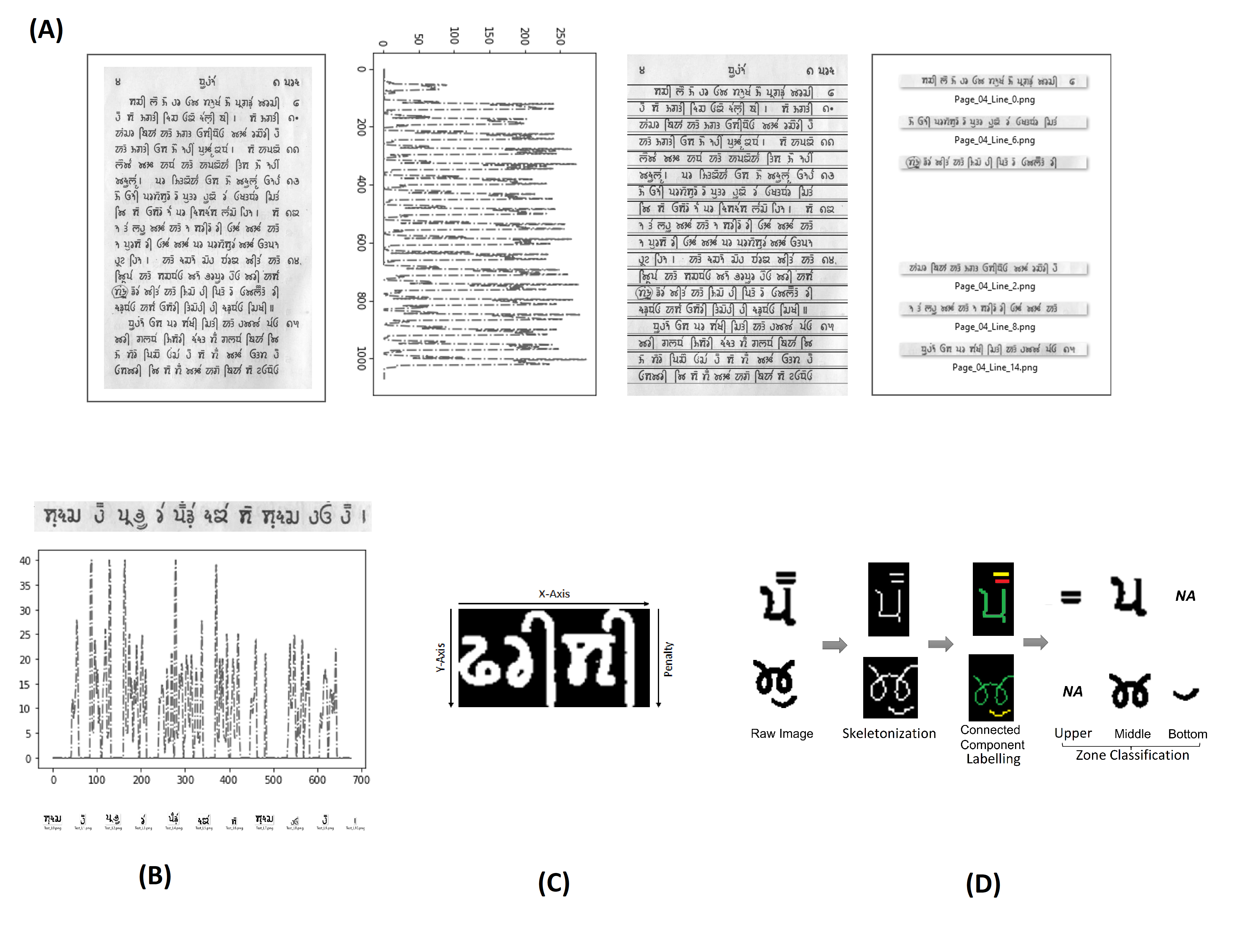}}
}
\end{figure*}
\begin{itemize}
    \setlength\itemsep{-0.35em}
\item A - Segmentation of page into lines.
\item B - Segmentation of line into words.
\item C - Segmentation of word into characters.
\item D - Segmentation of character into symbols.
\end{itemize}

\newpage
\label{appendixG}
\section*{Appendix G: Baseline studies.}
\textbf{TABLE A -}

\begin{flushleft}
\begin{longtable} [l]{| p{.15\textwidth} | p{.25\textwidth} |} 
 \hline
\textbf{Script Name} & \textbf{Annotated Symbols} \\[0.5ex]
\hline
 {Devanagiri} & {121K}\\ 
 {Gurumukhi} &  {111K}\\ 
 {Gujarati} & {121K}\\ 
 {Oriya} &  {115K}\\ 
 {Bangali} &  {122K}\\ 
 {Tamil} & {116K}\\ 
 \hline
\end{longtable}
\end{flushleft}

The annotated dataset is taken from \cite{akshara} which is the largest corpus of annotated Indic scripts. We have limited the size of the corpus to approx (110-120)K symbols to have all scripts in similar size. We used this initial dataset to build a base model for that script and later fine tune it for Takri.
\\
\\
\\

\textbf{TABLE B -}
\begin{longtable} {| p{.15\textwidth} | p{.45\textwidth} | p{.3\textwidth}|} 
 \hline
\textbf{Script Name} & \textbf{SOTA Accuracy on base script} & \textbf{Accuracy on Takri with 100K labelled dataset} \\[0.5ex]
\hline
 {Devanagiri} & {97.9\%} \cite{devBeng} & {86\%}\\ 
 {Gurumukhi} & {96\%} \cite{gurumukhi} & {86\%}\\ 
 {Gujarati} & {96\%} \cite{gujarati} & {87\%}\\ 
 {Oriya} & {96\%} \cite{oriya} & {87\%}\\ 
 {Bangali} &  {97\%} \cite{devBeng} & {85\%}\\ 
 {Tamil} & {(94-97)\%} \cite{tamil} & {78\%}\\ 
 \hline
\end{longtable}
We used our baseline models created using annotated symbol (from Table A) and used 100K annotated Takri tokens to fine tune for our purpose. SOTA accuracy (in second column) is for base script (as mentioned in column one) and Accuracy on Takri (in last column) is for Takri script.
\end{document}